\documentclass[letterpaper, 10 pt, journal, twoside]{IEEEtran}

\ifCLASSINFOpdf
\else
\fi

\usepackage{color}

\usepackage{graphics} 
\usepackage{amsmath,amssymb,amsfonts} 
\usepackage{bm}
\usepackage{subcaption}	 
\usepackage{graphicx}
\usepackage{multirow}
\usepackage{array}

\usepackage{url}
\usepackage{hyperref}
\usepackage{xcolor}

\usepackage{amsmath}

\usepackage{booktabs}
\usepackage{multirow}
\usepackage{textcomp}
\usepackage{setspace}

\usepackage{pifont}

\newcommand{\bT}{\mathbf{T}}
\newcommand{\bp}{\mathbf{p}}

\newcommand{\ba}{\mathbf{a}}
\newcommand{\bomega}{\boldsymbol{\omega}}

\newcommand{\br}{\mathbf{r}}

\newcommand{\bA}{\mathbf{A}}
\newcommand{\bg}{\mathbf{g}}

\newcommand{\bb}{\mathbf{b}}

\newcommand{\bn}{\mathbf{n}}

\newcommand{\bR}{\mathbf{R}}

\newcommand{\be}{\mathbf{e}}
\newcommand{\bx}{\mathbf{x}}

\newcommand{\bPhi}{\boldsymbol{\Phi}}

\newcommand{\blambda}{\boldsymbol{\lambda}}
\newcommand{\bM}{\mathbf{M}}
\newcommand{\bu}{\mathbf{u}}

\newcommand{\dotbR}{\dot{\mathbf{R}}}

\usepackage{mathrsfs}

\newcommand{\ddotbp}{\ddot{\mathbf{p}}}

\DeclareMathOperator*{\Exp}{Exp}

\usepackage[backend=bibtex,natbib=true,style=numeric-comp,sorting=none,giveninits=true,maxbibnames=99,url=false,doi=false]{biblatex}
\begin{document}
\title{Coco-LIC: Continuous-Time Tightly-Coupled LiDAR-Inertial-Camera Odometry using Non-Uniform B-spline}

\author{Xiaolei Lang$^{1}$, Chao Chen$^{1}$, Kai Tang$^{1}$,
Yukai Ma$^{1}$,  Jiajun Lv$^{1}$,
Yong Liu$^{1,*}$,
Xingxing Zuo$^{2,*}$

\thanks{Manuscript received: June 20, 2023; Revised August 17, 2023; Accepted September 5, 2023.}
\thanks{This paper was recommended for publication by Editor Javier Civera upon evaluation of the Associate Editor and Reviewers’ comments. This work is partially supported by the National Natural Science Foundation of China under grant NSFC 61836015.}
\thanks{$^{1}$ The authors are with the Institute of Cyber-Systems and Control, Zhejiang University, Hangzhou, China.}%
\thanks{$^{2}$ The author is with the Department of Computer Engineering,
Technical University of Munich, Germany.}%
\thanks{$^*$ Xingxing Zuo and Yong Liu are the corresponding authors (Email: {\tt\small xingxing.zuo@tum.de; yongliu@iipc.zju.edu.cn}).}
\thanks{Digital Object Identifier (DOI): see top of this page.}
}
%
%

\markboth{IEEE Robotics and Automation Letters. Preprint Version. September, 2023}
{Lang \MakeLowercase{\textit{et al.}}: Coco-LIC: Continuous-Time LiDAR-Inertial-Camera Odometry using Non-Uniform B-spline} 

%



\maketitle

\begin{abstract}
In this paper, we propose an efficient continuous-time LiDAR-Inertial-Camera Odometry, utilizing non-uniform B-splines to tightly couple measurements from the LiDAR, IMU, and camera. 
In contrast to uniform B-spline-based continuous-time methods, our non-uniform B-spline approach offers significant advantages in terms of achieving real-time efficiency and high accuracy. This is accomplished by dynamically and adaptively placing control points, taking into account the varying dynamics of the motion. To enable efficient fusion of heterogeneous LiDAR-Inertial-Camera data within a short sliding-window optimization, we assign depth to visual pixels using corresponding map points from a global LiDAR map, and formulate frame-to-map reprojection factors for the associated pixels in the current image frame.
This way circumvents the necessity for depth optimization of visual pixels, which typically entails a lengthy sliding window with numerous control points for continuous-time trajectory estimation.
We conduct dedicated experiments on real-world datasets to demonstrate the advantage and efficacy of adopting non-uniform continuous-time trajectory representation. 
Our LiDAR-Inertial-Camera odometry system is also extensively evaluated on both challenging scenarios with sensor degenerations and large-scale scenarios, and has shown comparable or higher accuracy than the state-of-the-art methods. 
The codebase of this paper will also be open-sourced at \url{https://github.com/APRIL-ZJU/Coco-LIC}.
\end{abstract}

\begin{IEEEkeywords}
LiDAR-Inertial-Camera SLAM, Sensor Fusion, Localization, State Estimation
\end{IEEEkeywords}

%
\IEEEpeerreviewmaketitle

%
%
%
%

\section{INTRODUCTION}

\IEEEPARstart{A}{} wide range of sensors can be applied for accurate 6-DoF motion estimation, among which LiDAR, IMU, and camera might be the most popular and widely used. Due to the inherent complementarity of 
such three sensors, LiDAR-Inertial-Camera Odometry (LICO) has achieved higher robustness and accuracy than those which only utilize the component sensors, especially in challenging structure-less and texture-less environments, attracting significant attention~\cite{zhang2018laser, shao2019stereo, shan2021lvi, zhao2021super, zuo2019lic, zuo2020lic, wisth2021unified, lin2021r, lin2022r, zheng2022fast, lv2023continuous}. However, measurements from these multi-modal sensors are usually received at different rates and different time instants, which causes difficulty in fusing them in a discrete-time odometry system. A common solution is to interpolate the measurements or estimated poses at the same time instants from different sensors for fusion~\cite{qin2018vins,shan2020lio}. 

In contrast, the continuous-time trajectory representation can avoid the need for interpolation and naturally enables pose queries at any given time, thereby facilitating the fusion of high-rate, multi-rate, and asynchronous measurements from various sensors~\cite{lovegrove2013spline}.
Most of the existing LICO systems adopt discrete-time trajectory~\cite{zuo2019lic, shan2021lvi, lin2022r}, while there are some works adopt continuous trajectory representation~\cite{lv2023continuous}.

In terms of parameterizing 6-DoF continuous-time trajectories, most existing  methods adopt a cubic, cumulative B-spline which offers local control property~\cite{hug2020hyperslam, lovegrove2013spline, lv2021clins, lv2023continuous, lang2022ctrl}. In these approaches, control points of B-splines are often distributed uniformly. It has been proved that the spacing of control points greatly influences the accuracy and time performance of trajectory estimation~\cite{cioffi2022continuous}. 
In the case of uniform B-splines, the spacing of control points is typically predetermined, which determines the complexity of the trajectory. However, in many applications, there is no prior knowledge about the complexity of the trajectory to be estimated. Consequently, due to the inability to dynamically adjust the distribution of control points, uniform B-splines are prone to under- or over-parameterization~\cite{hug2020hyperslam}. An appealing alternative is the use of non-uniform B-splines, which allows for dynamic control point distribution instead of a fixed frequency~\cite{hug2020hyperslam, vandeportaele2017pose, oth2013rolling}.

In terms of fusion methods in a LIC system, the existing methods can be classified into two categories: loosely-coupled methods, which utilize the visual-inertial system to provide initial values for LiDAR scan matching ~\cite{zhang2018laser, shao2019stereo}, and tightly-coupled methods, which jointly optimize LiDAR measurements with visual or inertial data~\cite{shan2021lvi,zuo2020lic,lv2023continuous,lin2021r,lin2022r,zheng2022fast}.
Generally, tightly-coupled methods tend to require higher computation to fuse raw measurements from various sensors but appear more accurate and robust. To formulate effective constraints for the estimated states by the LiDAR and the camera, data associations are required. 
For the LiDAR, edge and planar features~\cite{zhang2014loam} extracted based on curvature are commonly adopted. The extracted features from different scans are associated by finding closet neighbors~\cite{zhang2014loam,zuo2020lic}. For the camera, extracted visual features from images are associated by optical flow tracking~\cite{lucas1981iterative} or distinctive descriptors such as ORB~\cite{rublee2011orb}. Beyond intra-sensor data associations, there are also associations across sensors such as initializing or assigning the depth of visual features or pixels~\cite{zhang2014real,zheng2022fast}.
Our previous work CLIC~\cite{lv2023continuous} tightly couple LIC data within the  continuous-time trajectory optimization based on uniform B-splines. 
However, it disregards cross-sensor associations and instead relies on triangulating visual features using a lengthy sliding window of visual keyframes. This approach unavoidably involves numerous control points and adversely affects runtime efficiency of continuous-time framework.

In this work, we develop a continuous-time tightly-coupled LICO system based on the non-uniform B-spline trajectory representation, where control points are dynamically distributed and LIC data are tightly and efficiently coupled. Briefly, the contributions are summarized as follows: 
\begin{itemize}
\item We propose a LiDAR-Inertial-Camera Odometry system termed as Coco-LIC, utilizing continuous-time trajectory parameterized by non-uniform B-splines. Compared to methods based on uniform B-splines, control points here are dynamically placed through our proposed simple but effective distribution strategy.  
\item
We naturally fuse LIC data without any interpolation. Based on the reconstructed LiDAR point cloud map and optical-flow tracking of visual pixels, we formulate frame-to-map reprojection errors for the current image frame, excluding the depth estimation and optimization for visual pixels. This couples LiDAR and camera data in a tightly-coupled and highly-efficient way.
\item We specifically verify the necessity of the non-uniform placement of control points in real-world scenarios, and prove the efficacy of our control point spacing strategy. Furthermore, the entire LICO system is extensively tested on several challenging datasets, demonstrating its real-time performance and high accuracy.
\end{itemize} 
\section{RELATED WORK}

\subsection{Continuous-Time SLAM}
Furgale et al.~\cite{furgale2012continuous, furgale2015continuous} are among the first to present a full probabilistic derivation of the continuous-time state estimation for solving the SLAM problem based on B-splines. Afterward, they enable the joint estimation of the spatial and temporal extrinsic between different sensors by continuous-time batch optimization, which has been validated on the Visual-Inertial system and stereo LIC system~\cite{furgale2013unified, rehder2016general}. Lovegrove et al.~\cite{lovegrove2013spline} utilize the continuous-time formulation to incorporate measurements from rolling shutter cameras and IMU. As a well-crafted and comprehensive continuous-time calibration package for the LiDAR-Inertial system, Lv et al.~\cite{lv2022observability} simultaneously estimate the intrinsic and extrinsic while addressing degenerated motions. Different from these offline and batch calibration tasks, some works start to apply the continuous-time representation to online incrementally estimate odometry for the LiDAR-Inertial system~\cite{lv2021clins, park2021elasticity}, multi-camera systems~\cite{yang2021asynchronous}, event cameras~\cite{mueggler2018continuous}, rolling shutter cameras~\cite{lang2022ctrl} and so on. With the finding of the efficient derivative computation for cumulative B-splines on Lie groups~\cite{sommer2020efficient}, continuous-time odometry tends to achieve sub-real-time or even real-time performance~\cite{lv2023continuous}. A dedicated marginalization strategy in slid-window estimators is also proposed for the continuous-time framework~\cite{lang2022ctrl, lv2023continuous}. 

Continuous-time formulation benefits the fusion of high-rate, multi-rate, and asynchronous sensors. Cioffi et al.~\cite{cioffi2022continuous} further compare the continuous-time methods with the discrete-time methods comprehensively and find the frequency of the uniformly distributed control points in B-splines matters a lot for the trajectory estimation. In real practice, the smoothness of a trajectory can vary significantly, and each segment actually requires a different density of control points to accurately model the real trajectory~\cite{anderson2014hierarchical}. An adaptive metric for spacing control points is proposed in~\cite{oth2013rolling} by comparing the objective function cost against its expected value. Anderson et al.~\cite{anderson2014hierarchical} then adopt this scheme in continuous-time pose-graph optimization and find that sharp variations in robot velocity usually indicate the requirement for more control points. Vandeportaele et al.~\cite{vandeportaele2017pose} add more control points where the angular velocity or linear acceleration is larger, and vice versa. 
Hug et al.~\cite{hug2020hyperslam} recommend employing a more generic and non-uniform representation to prevent the under or over-parameterization. Admittedly, although the non-uniform formulation is critical for continuous-time trajectory estimation, it has only been employed in just a few batch offline applications, lacking implementation and verification in online odometry systems. Our work, to the best of our knowledge, is among the \textbf{\textit{first}} to adopt non-uniform continuous-time representation in the odometry task and demonstrates its efficacy through extensive experiments.

\subsection{LiDAR-Inertial-Camera Fusion for SLAM}
As early works of LICO, \cite{zhang2018laser, shao2019stereo} leverage monocular or stereo VIO to initialize LiDAR scan matching in a loosely-coupled way. Built upon a factor graph and composed of two subsystems, a LIO and a VIO, LVI-SAM~\cite{shan2021lvi} shows robustness in various LiDAR-degenerated or visual-failure scenarios.  Similarly, characterized by an IMU-centric estimator, Super-Odometry~\cite{zhao2021super} works well with an extra IMU odometry subsystem. Yet, the following tightly-coupled methods maintain only one state vector and show brilliant performance. LIC-fusion~\cite{zuo2019lic, zuo2020lic} and CLIC~\cite{lv2023continuous} both tightly fuse LiDAR features, sparse visual features, and IMU measurements within a sliding window, while the former adopts MSCKF~\cite{mourikis2007multi} filter based framework~\cite{geneva2020openvins} and the latter is based on continuous-time optimization framework. Wisth et al.~\cite{wisth2021unified} develop a unified multi-modal landmark tracking method and fuse IMU measurements with visual and LiDAR landmarks. R2LIVE~\cite{lin2021r} executes Error-State Iterated Kalman Filter (ESIKF) update whenever a LiDAR scan or a camera image comes and continues to optimize camera poses and visual landmarks in the window.
Furthermore, R3LIVE~\cite{lin2022r} and FAST-LIVO~\cite{zheng2022fast} maximally reuse the map points selected from the global LiDAR map to conduct frame-to-map photometric updates in an ESIKF framework, which avoids the triangulation and optimization for visual features over  multiple keyframes.

Unlike R3LIVE~\cite{lin2022r} and FAST-LIVO~\cite{zheng2022fast}, our method diverges by simultaneously fusing data from three sensors within a fixed time interval, rather than fusing LiDAR-Inertial or Visual-Inertial data separately.

\section{METHODOLOGY}
\label{sec:method}

\begin{figure}[t]
    \centering
    \includegraphics[width=0.85\linewidth]{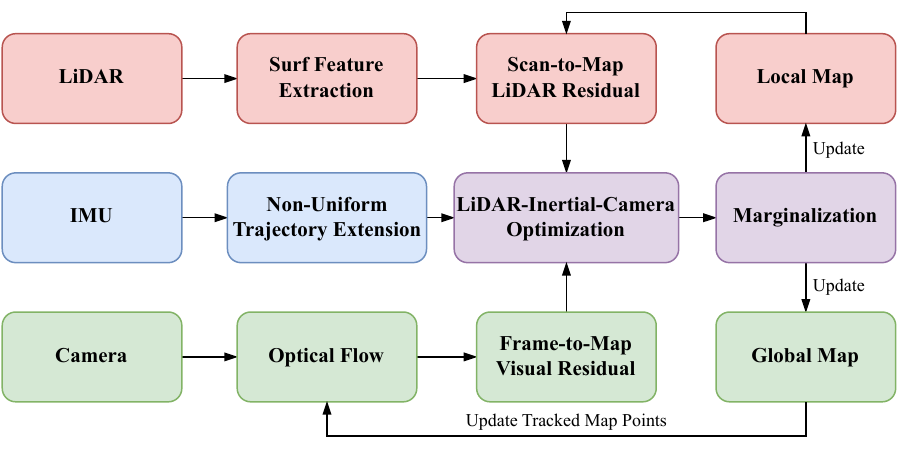}
    \caption{The pipeline of Coco-LIC.}
    \label{fig:sys_pipeline}
    \vspace{-1.0em}
\end{figure}

We first introduce the convention in this paper. We denote the 6-DoF rigid transformation by ${}^A_B\bT \in SE(3) \in \mathbb{R}^{4\times4}$, which transforms the point ${}^B\bp$ in the frame $\{B\}$ into the frame $\{A\}$. ${}^A_B\bT=\begin{bmatrix} {}^A_B\bR & {}^A\bp_B \\ \mathbf{0} & 1 \end{bmatrix}$ 
consists of rotation ${}^A_B\bR \in SO(3)$ and translation ${}^A\bp_B \in \mathbb{R}^{3}$. Exp$(\cdot)$ maps Lie Algebra to Lie Group and Log$(\cdot)$ is its inverse operation. $(\cdot)_{\wedge}$ maps a 3D vector to the corresponding skew-symmetric matrix, while $(\cdot)_{\vee}$ is its inverse operation.
The control points in the time interval $\left[t_a, t_{b}\right)$ are denoted as $\mathbf{\Phi}(t_a, t_{b})$.

\subsection{System Overview}
Fig.~\ref{fig:sys_pipeline} shows the pipeline of Coco-LIC. 
In the beginning, we assume the system is stationary and initialize the IMU bias and gravity by IMU measurements similar to~\cite{geneva2020openvins}. From now on, the continuous-time trajectory is extended and optimized every $\Delta t$ (0.1 in this letter) seconds based on the frequency of LiDAR. Suppose we have estimated the trajectory before $t_{\kappa-1}$, 
we will then estimate the trajectory in $[t_{\kappa-1}, t_{\kappa})$ once the LIC data in this time interval is ready, where $t_{\kappa} = t_{\kappa-1} + \Delta t$. We first dynamically decide the number of control points in the next $\Delta t$ seconds utilizing the proposed adaptive non-uniform technique (Sec.~\ref{sec:adaptive-technique}).
Subsequently, the newly added control points are initialized and further optimized in a manner of factor graph (Sec.~\ref{sec:lvi-opt}), using LiDAR planar points~\cite{zhang2014loam}, IMU raw measurements and the latest image in $[t_{\kappa-1}, t_{\kappa})$. Finally, after marginalization, we update the local and global LiDAR map for scan-to-map LiDAR association and frame-to-map visual association (Sec.~\ref{sec:tracked_map}), respectively.

In summary, we aim to estimate the following states by LIC data over $[t_{\kappa-1}, t_\kappa)$:
\begin{equation}
\begin{aligned}
    \label{eq:state}
    \mathcal{X}^{\kappa} &= \{\mathbf{\Phi}(t_{\kappa-1}, t_{\kappa}),\, \bx_{I_b}^{\kappa}\} , \\
	\bx_{I_b}^{\kappa} &= \{ \bb_{g}^{\kappa-1},\, \bb_{a}^{\kappa-1},\, \bb_{g}^{\kappa},\, \bb_{a}^{\kappa} \} ,
\end{aligned}
\end{equation}
which include control points $\mathbf{\Phi}(t_{\kappa-1}, t_{\kappa})$ and IMU bias $\bx_{I_b}^{\kappa}$. $\bb_g$ and $\bb_a$ indicate gyroscope bias and accelerometer bias, respectively. The IMU biases during $[t_{\kappa-1}, t_\kappa)$ are assumed to be $\bb_{g}^{\kappa-1}$ and $\bb_{a}^{\kappa-1}$ for simplicity. They are under Gaussian random walk and evolve to $\bb_{g}^{\kappa}$ and $\ba_{g}^{\kappa}$ at time instant $t_\kappa$. We denote all LiDAR planar points in $[t_{\kappa-1}, t_\kappa)$ as $\mathcal{L}_{\kappa}$, all IMU raw data as $\mathcal{I}_{\kappa}$, and the latest image frame as $\mathcal{F}_{\kappa}$.

\subsection{Non-Uniform Continuous-Time Trajectory Formulation}
\label{sec:continuous_time_trajectory}
In this paper, we adopt two non-uniform cumulative B-splines to separately parameterize the 3D rotation and the 3D translation. The 6-DoF poses at time $t\in [t_i, t_{i+1})$ of a continuous-time trajectory of degree $k$ are denoted by:
\begin{equation}
\begin{aligned}
    \bR(t) &=\bR_{i-k} \cdot  \prod_{j=1}^{k} \operatorname{Exp}\left(\lambda_{j}(t) \cdot \text{Log}\left(\mathbf{R}_{i-k+j-1}^{-1} \mathbf{R}_{i-k+j}\right)\right), \\
    \bp(t)&=\bp_{i-k}+\sum_{j=1}^{k} \lambda_{j}(t) \cdot \left(\bp_{i-k+j} - \bp_{i-k+j-1} \right), \\
    \blambda(t) &= \widetilde{\bM}^{(k+1)}(i) \bu, 
    \bu = {\begin{bmatrix} 1 \ u \ \cdots \ u^{k} \end{bmatrix}}^{\top},
    u = \frac{t-t_{i}}{t_{i+1}-t_i} ,
\label{eq:b-spline}
\end{aligned}
\end{equation}
where $\bR_x$ and $\bp_x$ denote control points ($x \in \{i-k, \cdots,i \}$)~\cite{Shifazhong1994ComputerAidedGeometricDesignandNURBS}. $t_{i-1}$ and $t_i$ represent any two adjacent knots of B-splines.~\cite{lowther2003teaching} The knots  $\{t_{0}, \, t_{1}, \, t_{2}, \,  \ldots\}$ of B-splines are non-uniformly distributed together with the control points, thus the cumulative blending matrix $\widetilde{\mathbf{M}}^{(k+1)}(i)$ is non-constant which is the main difference between uniform and non-uniform B-splines~\cite{qin1998general}. $\lambda_{j}(t)$ is an element of vector $\blambda(t)$ with index $j$. Cubic B-spline is adopted in this paper, which implies $k = 3$, and the blending matrix ${(\mathbf{M}^{(4)}(i)})^{\top}$ is derived as follows~\cite{qin1998general, sommer2020efficient}: 
\begin{align}
    & \left[\begin{array}{cccc}
m_{0,0} & 1-m_{0,0}-m_{0,2} & m_{0,2} & 0 \\
-3 m_{0,0} & 3 m_{0,0}-m_{1,2} & m_{1,2} & 0 \\
3 m_{0,0} & -3 m_{0,0}-m_{2,2} & m_{2,2} & 0 \\
-m_{0,0} & m_{0,0}-m_{3,2}-m_{3,3} & m_{3,2} & m_{3,3}
\end{array}\right],
\end{align}
$m_{a,b}$ denotes the element in row $a$, column $b$, computed by knots $t_{i-2} \sim t_{i+3}$ \cite{qin1998general}. $\widetilde{\mathbf{M}}^{(4)}(i)$ can be derived accordingly:
\begin{align}
\label{eq:cumu_blend}
\widetilde{\mathbf{M}}^{(k+1)}(i)_{m,n} = \sum_{s=m}^{k}\mathbf{M}^{(k+1)}(i)_{s,n}.
\end{align}

The continuous-time trajectory of IMU in the global frame $\{G\}$ is denoted as ${}^G_I\bT(t) = \left[{}^G_I\bR(t), {}^G\bp_I(t)\right]$. In this letter, the extrinsics between LiDAR/camera and IMU are pre-calibrated, 
so we can handily get LiDAR trajectory ${}^G_L\bT(t)$ and camera trajectory ${}^G_C\bT(t)$. B-splines provide closed-form analytical derivatives w.r.t. temporal variables~\cite{sommer2020efficient, furgale2013unified}, leading to easily computed angular velocity and linear acceleration. The time derivatives of the B-splines can be derived as:
\begin{align}
    \label{eq:spline_dR}
    {}_I^G\dotbR(t) &= 
    {}_I^G\bR(t) \cdot \left({}^I\bomega(t)\right)_{\wedge}\nonumber \\ 
    &=
    \bR_{i-k}\left(\dot{\bA}_1 \bA_2 \bA_3 + \bA_1 \dot{\bA}_2 \bA_3 + \bA_1 \bA_2 \dot{\bA}_3\right) \\
    \label{eq:spline_ddp}
    {}^G\ba(t) &= {}^G\ddotbp_I(t) = \sum_{j=1}^{k}{\ddot{\lambda}_j(t)\cdot \left(\bp_{i-k+j} - \bp_{i-k+j-1} \right)}  
\end{align}
where $\dot{\bA}_j = \Exp\left(\dot{\lambda}_j(t)\cdot \text{Log}\left(\mathbf{R}_{i-k+j-1}^{-1} \mathbf{R}_{i-k+j}\right) \right)$.

\begin{figure}[t]
    \centering
    \includegraphics[width=0.7\linewidth]{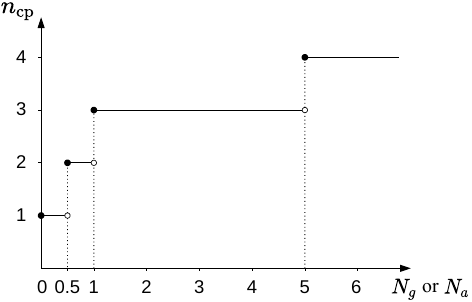}
    \caption{The number of control points to add depends on IMU. $N_g$ and $N_a$ might correspond to the different values of $n_{\text{cp}}$, while we choose the larger one.}
    \label{fig:cp-gear}
    \vspace{-1.5em}
\end{figure}

\subsection{Adaptive Non-Uniform Technique}
\label{sec:adaptive-technique}

New control points should be added to extend the trajectory, the number of which is set as a constant value in uniform-spline-based methods. However, such uniform placement might fail to precisely model the trajectory segments with violent motions, or introduce redundant control points where the motion is steady, leading to an increase of the
computational cost. To address the issue of the uniform B-spline not being able to flexibly accommodate trajectory complexity online, we employ an adaptive non-uniform technique analogous to~\cite{vandeportaele2017pose}.
Specifically, we try to get the norm of average angular velocity $N_{g}$ and linear acceleration $N_{a}$ in the global frame in $[t_{\kappa-1}, t_\kappa)$:
\begin{equation}
\begin{aligned}
    N_{g} &= \frac{1}{n}\left\lVert \sum_{i}^{n} {}^G_I{\bR_{m_i}} {}^I{\bomega_{m_i}}\right\rVert, \\
    N_{a} &= \frac{1}{n}\left\lVert  \sum_{i}^{n}
    \left(
    {}^G_I{\bR_{m_i}} {}^I{\ba_{m_i}} - {}^G\bg
    \right)
    \right\rVert,
\end{aligned}    
\end{equation}
where ${}^I{\bomega_{m_i}}$ and ${}^I{\ba_{m_i}}$ are angular velocity and linear acceleration measured by IMU, respectively. $n$ denotes the number of IMU measurements. The rotation ${}^G_I{\bR_{m_i}}$ is integrated from $t_{\kappa-1}$ using raw IMU measurements. ${}^G\bg = \begin{bmatrix} 0&0&9.8\end{bmatrix}^T$ is the gravity in the global frame. 
Both $N_g$ and $N_a$ reflect the intensity of the motion. We then determine the number $n_{\text{cp}}$ of control points to add according to Fig.~\ref{fig:cp-gear}, which is proven to be effective and generic in our experiments. The new control points are all initially assigned with the value of the last control point in the latest optimization. They are uniformly distributed in $[t_{\kappa-1}, t_\kappa)$ with the time interval divided uniformly by $\frac{1}{n_{\text{cp}}}$. Then we can query the pose at any time instant over $[t_{\kappa-1}, t_\kappa)$. 

Subsequently, the newly added control points are further optimized by solving a factor graph optimization as Eq.\eqref{eq:lvio-graph} with only IMU factors and a prior factor specified in Sec.~\ref{sec:lvi-opt}, where we only optimize $\mathbf{\Phi}(t_{\kappa-1}, t_{\kappa})$ and keep the other states constant.

\subsection{LiDAR-Inertial-Camera Optimization}\label{sec:lvi-opt}
\subsubsection{Scan-to-Map LiDAR Factor}
In the continuous-time framework, we estimate the trajectory of a whole scan instead of merely estimating a single pose, which enables simultaneous motion distortion removal and trajectory estimation~\cite{lv2021clins}. Consider a LiDAR planar point ${}^{L}\bp$ in $\mathcal{L}_k$ measured at time $t$, we can transform it to the global frame by 
\begin{align}
    {}^{G}\hat{\bp} = {}^G_L\bR(t) {}^L\bp + {}^G\bp_L(t).
\end{align}
We find for ${}^{G}\hat{\bp}$ five closest neighbor points in the local LiDAR map, and fit a 3D plane using these neighbor points. The local LiDAR map is constructed atop keyscans that are selected by time and space~\cite{lv2021clins}.
Thus the point-to-plane error can be defined as:
\begin{align}
    \br_{L} = {}^{G}\bn^\top_{\pi} {}^{G}\hat{\bp} + {}^{G}d_{\pi},
\end{align}
where ${}^{G}\bn_{\pi}$ and ${}^{G}d_{\pi}$ denote the unit normal vector and the distance of the plane to the origin respectively. Jacobians of $\br_{L}$ w.r.t. control points can be found in~\cite{lv2023continuous}, the main difference lies in $\blambda(t)$ where the cumulative blending matrix is not constant according to Eq.\eqref{eq:b-spline}.

\begin{figure}[t]
    \centering
    \includegraphics[width=0.5\linewidth]{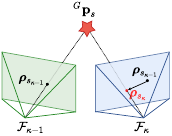}
    \caption{Projection of a LiDAR map point across image frames. {{}${}^G\bp_{s}$ in the global LiDAR map is fixed during optimization.}}
    \label{fig:reprojection}
    \vspace{-1.5em}
\end{figure}

\subsubsection{Frame-to-Map Visual Factor}
Similar to~\cite{lin2022r, zheng2022fast}, we associate 3D map points in the global LiDAR map (detailed in Sec.~\ref{sec:tracked_map}) with the image by projection. Suppose We have associated the map points $\mathcal{P}=\left\{{}^G\bp_{1}, \ldots, {}^G\bp_{m}\right\}$ in the last image frame $\mathcal{F}_{\kappa-1}$ with the corresponding pixels $\hat{\varrho}_{\kappa-1} = \left\{\hat{\boldsymbol{\rho}}_{1_{\kappa-1}}, \ldots, \hat{\boldsymbol{\rho}}_{m_{\kappa-1}}\right\}$. We then track these pixels $\hat{\varrho}_{\kappa-1}$ into frame $\mathcal{F}_{k}$ by KLT sparse optical flow~\cite{lucas1981iterative} and get tracked pixels $\varrho_{\kappa} = \left\{\boldsymbol{\rho}_{1_{\kappa}}, \ldots, \boldsymbol{\rho}_{m_{\kappa}}\right\}$. Afterward, RANSAC based on the fundamental matrix and PnP are successively leveraged to remove outlier associations, and we finally get a group of map points $\mathcal{P}$ successfully associated with pixels in the current image frame $\mathcal{F}_{k}$. Consider a map point ${}^G\bp_{s}$ observed in frame $\mathcal{F}_{k}$ with the optical-flow tracked pixel  $\boldsymbol{\rho}_{s_{k}} = \begin{bmatrix} u_s & v_s \end{bmatrix}^\top$. As shown in Fig.~\ref{fig:reprojection}, the reprojection error of the tracked LiDAR point in frame $\mathcal{F}_{k}$ is defined as:
\begin{align}
    \br_{C} = \pi_c\left( \frac{{}^C\hat{\bp}_{s}}{\be_{3}^{\top} {}^C\hat{\bp}_{s}} \right) - \begin{bmatrix} u_s \\ v_s \end{bmatrix}, \quad 
    {}^C\hat{\bp}_{s} = {}^G_C\bT^\top(t) {}^G\bp_{s},
\end{align}
where $\be_{i}$ is a $3\times1$ vector with its $i$-th element to be 1 and the others to be 0. $\pi_c(\cdot)$ denotes the projection function which transforms a point on the normalized image plane to a pixel. Note that we also apply Cauchy robust kernel to the reprojection error in optimization to further suppress outliers. In such a fashion based on the optical flow tracking of existing map points with known depth, we can avoid triangulation and sliding window optimization of visual features, thus keeping a short sliding window within $[t_{\kappa-1}, t_\kappa)$ for high efficiency and accuracy, without numerous control points involved.

\subsubsection{IMU Factor and Bias Factor}
We seamlessly use raw IMU measurements to formulate IMU factors like~\cite{lang2022ctrl,lv2023continuous}, instead of using IMU preintegration. For IMU data in $\mathcal{I}_\kappa$ at time $t$, we define the following IMU factor:
\begin{align}
    & \br_{I} = \begin{bmatrix} 
    {}^I\bomega(t) - {}^I\bomega_m + \bb_{g}^{\kappa-1} \\
    {}^I\ba(t) - {}^I\ba_m + \bb_a^{\kappa-1}
    \end{bmatrix},
\end{align}
and bias factor based on the random walk process:
\begin{align}
    \br_{I_b} = \begin{bmatrix} 
    \bb_{g}^{\kappa} - \bb_{g}^{\kappa-1} \\
    \bb_{a}^{\kappa} - \bb_{a}^{\kappa-1}
    \end{bmatrix},
\end{align}
where ${}^I{\boldsymbol{\omega}_m}, 
{}^I{\ba_m}$ are the raw measurements of angular velocity and linear acceleration at time $t$, respectively.
${}^I\ba(t)$ and ${}^I\bomega(t)$  can be computed from the derivatives of the continuous-time trajectory in Eqs.\eqref{eq:spline_dR} and \eqref{eq:spline_ddp} by:
\begin{align}
    {}^I\ba(t) & =
    {}_I^G\bR^\top(t) \left( {}^G\ba(t) - {}^G\bg \right), \notag \\
    {}^I\bomega(t) & = \left({}_I^G\bR^\top(t) \cdot {}_I^G\dotbR(t)\right)_{\vee}, \notag
\end{align} 
where the computation of ${}^I\bomega(t)$ has been further sped up using recurrence relation~\cite{sommer2020efficient} without costly computing ${}_I^G\dotbR(t)$.

\begin{figure}[t]
    \centering
    \includegraphics[width=\linewidth]{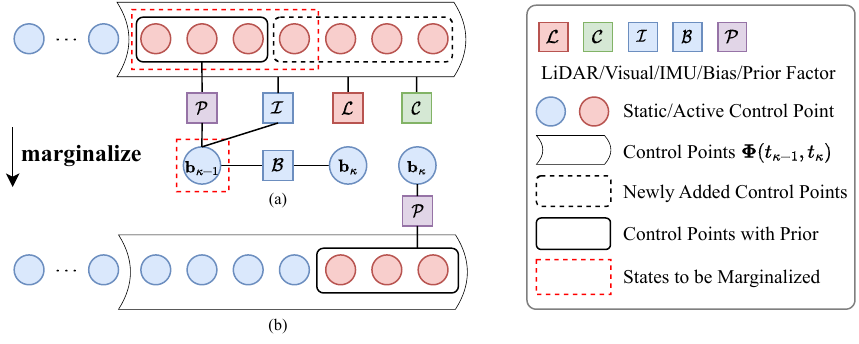}
    \caption{(a) Factor graph. (b) Marginalization.}
    \label{fig:graph-marg}
    \vspace{-1.5em}
\end{figure}
    
\subsubsection{Optimization and Marginalization}
To estimate the aforementioned states $\mathcal{X}^{\kappa}$, we jointly fuse LIC data in a factor graph as displayed in Fig.~\ref{fig:graph-marg} and formulate the following nonlinear least-squares problem:
\begin{equation}
\begin{aligned}
    \label{eq:lvio-graph}
    \arg \min_{\mathcal{X}^{\kappa}}
    \biggl\{ & \sum \left\|\br_L\right\|_{\Sigma_{L}}^{2}+
    \sum \left\|\br_C\right\|_{\Sigma_{C}}^{2}+
    \sum \left\|\br_{I}\right\|_{\Sigma_{I}}^{2}+   \\
    & 
    \sum \left\|\br_{I_b}\right\|_{\Sigma_{I_b}}^{2}+
    \sum \left\|\br_{\text{prior}}\right\|_{\Sigma_{\text{prior}}}^{2} \biggr\}
    \,,
\end{aligned}
\end{equation}
which is solved by the Levenberg-Marquardt algorithm in Ceres Solver~\cite{ceres-solver} and sped up by analytical derivatives. $\br_{\text{prior}}$ is the prior factor from marginalization. $\Sigma_{L}$, $\Sigma_{C}$, $\Sigma_{I}$, $\Sigma_{I_b}$, $\Sigma_{\text{prior}}$ are the corresponding covariance matrices.

After the optimization is done for the newly added control points and IMU bias, we marginalize states that will not be involved in the next $\Delta t$ seconds, that is $[t_{\kappa}, t_{\kappa+1})$, and obtain the prior factor which constrains the states as follows:
\begin{align}
    \mathcal{X}_{prior}^{\kappa} = \left\{ \bPhi(t_{\kappa-1}, t_{\kappa}) \cap \bPhi(t_{\kappa}, t_{\kappa+1}), \, \bb_{g}^{\kappa}, \bb_{a}^{\kappa} \right\} \text{,} \notag
\end{align}
where $\bPhi(t_{\kappa-1}, t_{\kappa}) \cap \bPhi(t_{\kappa}, t_{\kappa+1})$ represents the control points shared by the trajectory segments in $[t_{\kappa-1}, t_{\kappa})$ and $[t_{\kappa}, t_{\kappa+1})$. 

\subsection{LiDAR Map for Visual Factor}
\label{sec:tracked_map}

We maintain a global LiDAR map stored in voxels~\cite{lin2022r, zheng2022fast} for the fast query of map points associated with images. The voxel resolution is 0.1m in our experiments. After the optimization of the trajectory in $[t_{\kappa-1}, t_{\kappa})$, the global LiDAR map is updated with LiDAR planar points from $\mathcal{L}_{\kappa}$. For high-quality association, tracked map points in $\mathcal{P}$ with large reprojection error on $\mathcal{F}_{\kappa}$ are removed. We then project map points in the current FoV onto $\mathcal{F}_{\kappa}$ and add new successfully associated map points into $\mathcal{P}$. We ensure tracked map points are evenly distributed on images and prioritize retaining map points closer to the camera to alleviate the occlusion problem.


\section{EXPERIMENTS} \label{sec:exp}

In the experiments, we first compare and analyze the trajectory estimation accuracy and the time cost of uniform and non-uniform methods based on LiDAR-Inertial data.
Second, we evaluate the proposed Coco-LIC in degraded sequences and compare it with several typical open-source LIO, VIO, and LICO methods.
Finally, we conduct experiments on a large-scale dataset to further assess accuracy and time consumption in outdoor environments.

All experiments are executed on a desktop PC with an Intel i7-8700 CPU @ 3.2GHz and 32GB RAM. 
In one dataset, we keep the same parameters of Coco-LIC for all sequences for a fair comparison. We run all experiments six times and take the average values as results.
The noise parameters of the IMU are taken from the datasheet. Also, the estimated poses of Coco-LIC used for evaluation by evo~\cite{grupp2017evo} are queried from the continuous time trajectory at 100 Hz.

\begin{table}[t]
\caption{Description of the sequences.}
\resizebox{\linewidth}{!}{
\begin{tabular}{@{}cccc@{}}
\toprule
Seq & \begin{tabular}[c]{@{}c@{}}Duration\\ (second)\end{tabular} & \begin{tabular}[c]{@{}c@{}}Length\\ (m)\end{tabular} & Description \\ \midrule
\textit{Smooth1} & 78 & 19 & move gently in a square shape \\
\textit{Smooth2} & 117 & 17 & move gently in a butterfly shape \\
\textit{Smooth3} & 154 & 23 & move gently in a circle shape \\
\textit{Violent1} & 74 & 20 & large angular velocity \\
\textit{Violent2} & 91 & 115 & large linear acceleration \\
\textit{Violent3} & 121 & 78 & large angular velocity and linear acceleration \\
\textit{Hybrid1} & 91 & 37 &  smooth-violent-smooth \\
\textit{Hybrid2} & 137 & 69 & smooth-violent-smooth \\
\textit{Hybrid3} & 104 & 64 & smooth-violent-smooth-violent-smooth \\ \bottomrule
\end{tabular}
}
\label{tab:data-detail}
\vspace{-1.5em}
\end{table}

\begin{table*}[t]
\captionsetup{font={small}}
\caption{ The RMSE of APE results and the time consumption for optimization of LIO with different control point distributions (unit: meters/milliseconds). The best results are marked in bold.
Here, \textit{uni-x} means x number of control points per $\Delta t$ seconds, while \textit{non-uni} means dynamically placing control points by the adaptive technique specified in Sec.\ref{sec:adaptive-technique}. \textit{fail} means large RMSE over 1 m.}
\label{tab:ape-lio}
\centering
\resizebox{\linewidth}{!}{
\begin{tabular}{@{}cccccccccc@{}}
\toprule
 & \textit{Smooth1} & \textit{Smooth2}     & \textit{Smooth3} & \textit{Violent1} & \textit{Violent2} & \textit{Violent3} & \textit{Hybrid1} & \textit{Hybrid2} & \textit{Hybrid3} 
\\ \midrule
\textit{uni-1}                & 0.011 / \textbf{16.71}               & \textbf{0.027} / 19.12                     & 0.018 / 16.81                 & 0.355 / \textbf{17.32}                  & fail                   & fail                  & fail                  & fail                  & fail                  \\
\textit{uni-2}                & 0.012 / 24.06                & 0.048 / 28.50                     & 0.018 / 24.35                 & 0.060 / 24.84                  & 0.079 / \textbf{22.92}                  & 0.140 / \textbf{23.59}                  & 0.164 / 23.93                 & 0.059 / 24.70                 & 0.038 / 25.68                 \\
\textit{uni-3}                & 0.012 / 29.76                & 0.050 / 35.07                     & 0.018 / 29.48                 & 0.054 / 30.60                  & 0.052 / 28.64                  & 0.112 / 29.95                  & 0.105 / 25.09                 & 0.059 / 30.75                 & 0.035 / 32.49                 \\
\textit{uni-4}                & 0.012 / 31.48                & 0.048 / 38.83                     & 0.018 / 31.05                 & 0.052 / 34.05                  & 0.051 / 31.54                  & 0.102 / 33.45                  & 0.101 / 32.05                 & 0.060 / 34.54                 & 0.035 / 36.79                 \\
\textit{uni-5}                & 0.012 / 32.00                & 0.051 / 40.29 & 0.019 / 31.08                 & 0.054 / 35.33                  & 0.051 / 35.37                  & 0.104 / 34.98                  & 0.107 / 32.71                 & 0.060 / 36.39                 & 0.036 / 38.93                 \\
\textit{uni-8}                & 0.013 / 34.07                & 0.052 / 42.94                     & 0.018 / 32.87                 & 0.052 / 37.28                  & 0.060 / 37.14                  & 0.118 / 36.91                  & 0.113 / 34.84                 & fail                  & 0.037 / 41.91                 \\
\textit{uni-16}               & 0.049 / 38.80                & fail                      & fail                  & fail                   & fail                   & fail                   & fail                  & fail                  & fail                  \\
\textit{non-uni}              & \textbf{0.011} / 16.84                & 0.028 / \textbf{19.06}                     & \textbf{0.018} / \textbf{16.78}                 & \textbf{0.038} / 22.56                  & \textbf{0.051} / 29.14                  & \textbf{0.100} / 27.45                  & \textbf{0.100} / \textbf{20.27}                 & \textbf{0.059} / \textbf{24.12}                 & \textbf{0.035} / \textbf{24.54}                 \\ \bottomrule
\end{tabular}
}
\vspace{-1em}
\end{table*}

\subsection{Comparison of Uniform and Non-Uniform Placement}
We collect LiDAR-Inertial data by a sensor rig comprising a 16-beam 3D LiDAR Velodyne VLP-16
\footnote{\url{https://velodyneLiDAR.com/vlp-16.html}} 
at 10 Hz and an Xsens MTi-300 IMU
\footnote{\url{https://www.xsens.com/hubfs/Downloads/Leaflets/MTi-300.pdf}} 
at 400 Hz, with ground-truth data at 120 Hz recorded by a motion capture system. The accuracy of the motion capture is at the millimeter level. Note that cameras experience motion blur under violent motions leading to the poor quality of visual data, while extremely intense movements are contained in this experiment, thus we here only use LiDAR-Inertial data, which is sufficient for non-uniform verification. In real-world applications, motion can be divided into three modes, that is,  \textit{smooth}, \textit{violent}, and \textit{hybrid}:
\begin{itemize}
    \item \textit{smooth} case: ground vehicles moving on smooth roads, and cleaning robots working in office park.
    \item \textit{violent} case: quadruped robots repeatedly impacting the ground, and aerial robots quickly avoiding obstacles.
    \item \textit{hybrid} case: humans holding a scanning device to build a map of the environment.
\end{itemize}
We gather data with a total of nine sequences detailed in Table~\ref{tab:data-detail}, including three motion patterns mentioned above.

Table~\ref{tab:ape-lio} summarizes the RMSE of APE and the time consumption for different control point distributions in LIO. 
Here, \textit{uni-x} means there is $x$ number of control points per $\Delta t$ seconds using uniform B-splines, while \textit{non-uni} represents dynamically placing control points by the adaptive non-uniform technique specified in Sec.~\ref{sec:adaptive-technique}. 

\noindent \textbf{1) Smooth sequences:} 
\textit{uni-1} achieves the highest accuracy, and increasing the number of control points can lead to slightly lower accuracy or even failure to estimate the trajectory (\textit{uni-16}), which suggests that fewer control points are sufficient to accurately model the trajectory with lower complexity for smooth motions, while too many control points might cause over-fitting. Additionally, the time consumption for optimization significantly increases with more control points, probably due to the increase in the dim of states. In contrast, \textit{non-uni} adaptively places control points at a density of 1 per $\Delta t$ seconds in most cases, obtaining comparable accuracy and computation time to \textit{uni-1}. 

\noindent \textbf{2) Violent sequences:} 
The trajectories exhibit high complexity under large angular velocity and linear acceleration, which \textit{uni-1} fails to adapt to, resulting in poor pose estimation. Increasing the number of control points can improve accuracy, but still, excess ones will decrease accuracy and increase computation time. On the other hand, \textit{non-uni} adjusts the density of control points based on inertial data at each time interval, achieving the optimal accuracy with moderate computation cost.

\begin{figure}[t]
    \centering
    \includegraphics[width=1.0\linewidth]{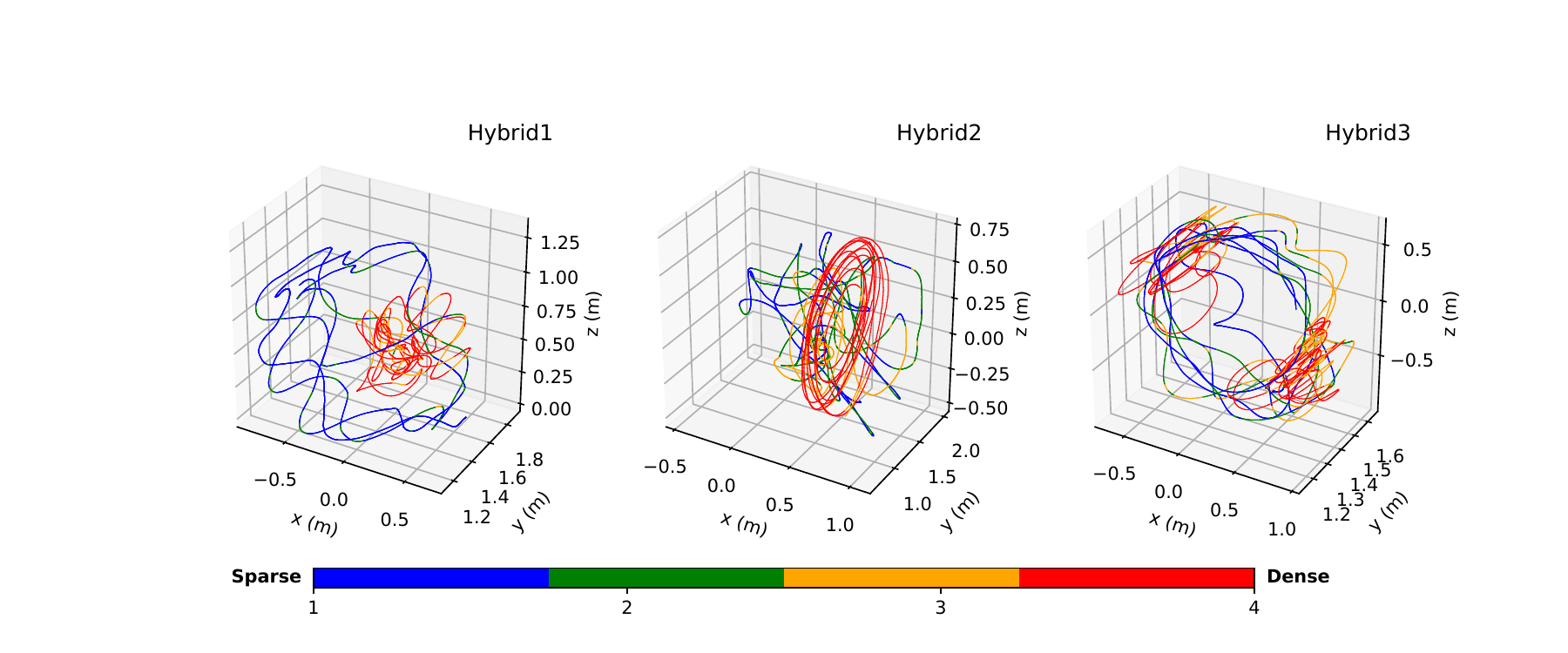}
    \captionsetup{font={small}}
    \caption{Trajectories estimated by LIO with non-uniform distribution of the control points on hybrid sequences. The different colors of the trajectories correspond to different densities of control points. From blue to red, control points change from sparse to dense.}
    \vspace{-1.0em}
    \label{fig:color-traj}
\end{figure}

\begin{figure}[t]
	\centering
    \includegraphics[width=\linewidth]{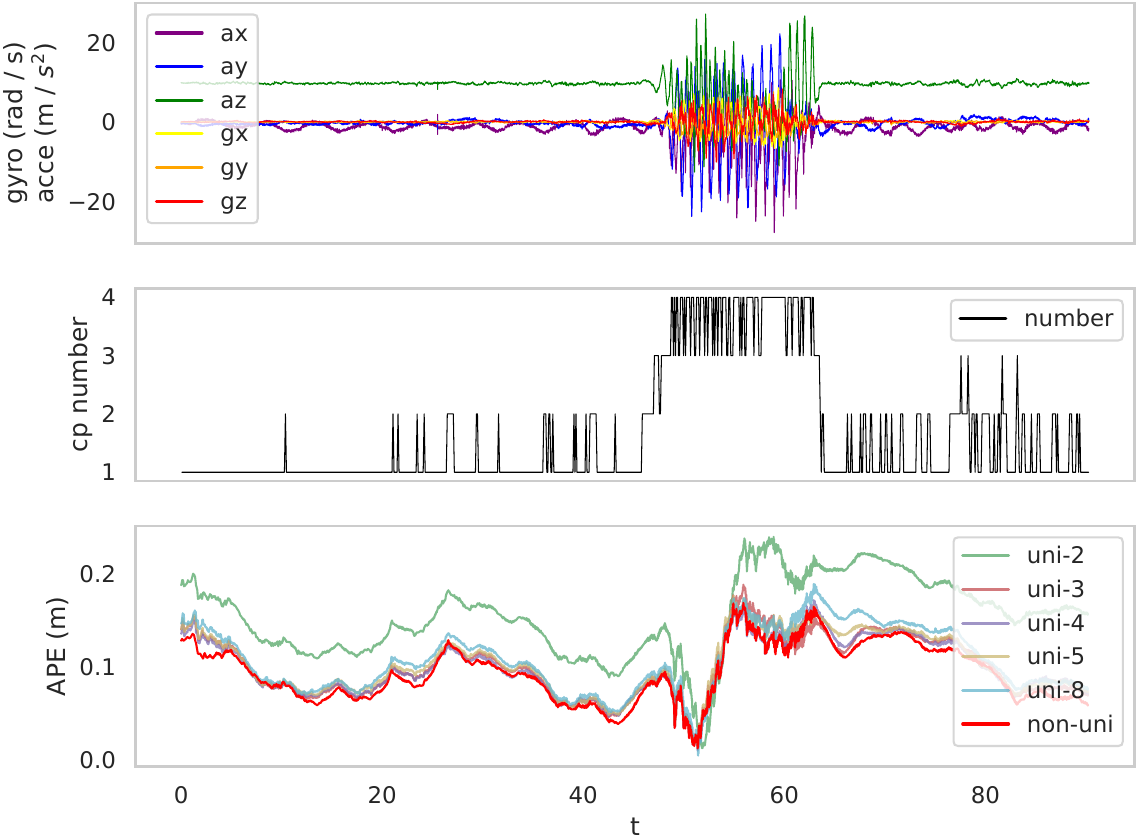}
	\captionsetup{font={small}}
	\caption{The top shows the IMU data in \textit{Hybrid1} and the middle shows the number of control points set per $\Delta t$ with time by adaptive technique. The bottom shows the pose error curve over time based on different distributions of the control points and the red one denotes the result of \textit{non-uni}.}
	\vspace{-1.5em}
	\label{fig:pose-error}
\end{figure}

\noindent \textbf{3) Hybrid sequences:} 
It is inadequate for \textit{uni-1} to accurately fit trajectory segments with intense motions, causing the failure of the estimation. Adding adequate control points can enhance the accuracy, whereas it appears to be unnecessary and incurs excess computation consumption for simple and smooth motion profiles. By adding control points during intense motions and reducing them during smooth motions, as displayed in Fig.~\ref{fig:color-traj}, our \textit{non-uni} attains the highest accuracy at almost the lowest time cost. Fig.~\ref{fig:pose-error} also depicts that \textit{non-uni}  maintains a small pose error throughout the whole trajectory owing to the adaptive non-uniform technique.

\begin{table*}[t]
\captionsetup{font={small}}
\caption{The start-to-end drift error on all challenging sequences of R3LIVE and FAST-LIVO (unit: meters/degrees). \textit{degenerate\_seq\_00}, \textit{degenerate\_seq\_01}, and \textit{degenerate\_seq\_02} are from R3LIVE. \textit{LiDAR\_Degenerate} and \textit{Visual\_Challenge} are from FAST-LIVO. The best results are marked in bold. \textit{fail} means the start-to-end drift error is larger than 10 meters.
}
\label{tab:start-end}
\centering
\resizebox{0.9\linewidth}{!}{
\begin{tabular}{@{}cccccccc@{}}
\toprule
\multicolumn{1}{l}{} & Degeneration Type${}^{(*)}$ & FAST-LIO2   & VINS-Mono   & R3LIVE      & FAST-LIVO   & CLIC        & Coco-LIC   \\ \midrule
\textit{degenerate\_seq\_00}  & L & 9.019 / 3.441 & 0.807 / 12.736 & 0.035 / \textbf{0.405} & 0.420 / 3.621 & 0.031 / 0.578 & \textbf{0.016} / 0.428 \\
\textit{degenerate\_seq\_01}  & L & 3.161 / 15.632 & 2.455 / 3.523 & 0.114 / 0.536 & 0.881 / 2.248 & 1.287 / 1.293 & \textbf{0.062} / \textbf{0.262} \\
\textit{degenerate\_seq\_02}  & L, V1 & fail & fail & \textbf{0.065} / \textbf{1.124} & 3.855 / 9.303 & 2.457 / 2.823 & 0.122 / 1.283 \\ 
\textit{LiDAR\_Degenerate}    & L & 0.798 / 1.816 & 1.644 / 2.572 & 8.733 / \textbf{1.861} & 0.049 / 2.025 & 6.811 / 5.911 & \textbf{0.045} / 2.497 \\
\textit{Visual\_Challenge}    & L, V2 & 2.957 / 2.707 & fail & 0.234 / 0.751 & \textbf{0.043} / \textbf{0.408} & 5.262 / 4.537 & 0.166 / 0.889 \\ \bottomrule
\multicolumn{7}{l}{${}^{(*)}$ L denotes LiDAR faces the ground or walls. V1, V2 denote camera faces a white wall and undergoes motion blur, resp.}
\\
\end{tabular}
}
\vspace{-1.0em}
\end{table*}

\begin{table}[t]
\centering
\captionsetup{font={small}}
\caption{The RMSE (m) of APE results on UrbanNav dataset. For simplicity, the sequences UrbanNav-HK-Medium-Urban-1, UrbanNav-HK-Deep-Urban-1, and UrbanNav-HK-Harsh-Urban-1 are referred to as \textit{Medium}, \textit{Deep}, and \textit{Harsh} respectively.}
\resizebox{0.6\linewidth}{!}{
\begin{tabular}{@{}cccc@{}}
    \toprule
    \multicolumn{1}{l}{} & \begin{tabular}[c]{@{}c@{}}\textit{Medium}\\ (3.64km)\end{tabular} & \begin{tabular}[c]{@{}c@{}}\textit{Deep}\\ (4.51km)\end{tabular}   & \begin{tabular}[c]{@{}c@{}}\textit{Harsh}\\ (3.04km)\end{tabular}  \\ \midrule
    FAST-LIO2 & 6.734  & 6.335  & 2.820   \\
    VINS-Mono & 102.582 & 63.894 & fail   \\
    LVI-SAM   & 7.456 & 8.607  & 25.684   \\
    FAST-LIVO & 7.331 & 7.159 & 2.787 \\
    CLIC      & 6.923  & 6.001   & 2.415   \\
    Coco-LIC & \textbf{6.031}  & \textbf{5.688}  & \textbf{2.189}
    \\ \bottomrule
\end{tabular}
}
\vspace{-1.3em}
\label{tab:lvi-ape}
\end{table}

\subsection{Robustness and Accuracy Evaluation of LiDAR-Inertial-Camera Odometry}
We test our LICO (Coco-LIC) on both challenging datasets and large-scale datasets, comparing it against four LIC methods: LVI-SAM~\cite{shan2021lvi}, R3LIVE~\cite{lin2022r}, FAST-LIVO~\cite{zheng2022fast}, CLIC~\cite{lv2023continuous} and a LIO system FAST-LIO2~\cite{xu2022fast}, as well as a VIO system VINS-Mono~\cite{qin2018vins}. Since LVI-SAM only supports rotating LiDAR and R3LIVE only supports solid-state LiDAR, Tab.~\ref{tab:start-end} misses the results of LVI-SAM and Tab.~\ref{tab:lvi-ape} misses R3LIVE. During the experiments, the loop closure module is disabled for pure odometry evaluation, and the reported results are the average of 6 runs. Note that, CLIC assigns three control points every $\Delta t$ seconds, as described in~\cite{lv2023continuous}, whereas Coco-LIC dynamically places control points.

\subsubsection{Challenging Degenerate Dataset}

\begin{figure}[t]
    \centering
    \includegraphics[width=1.0\linewidth]{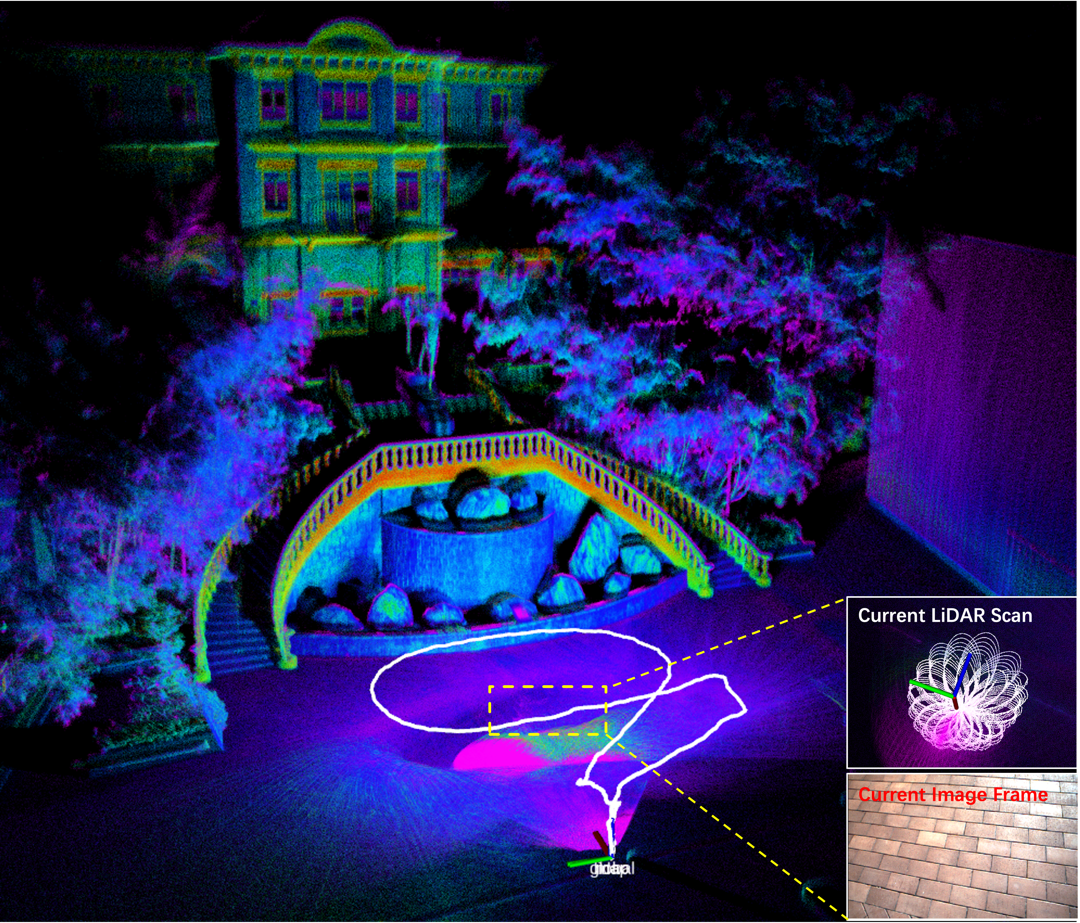}
    \captionsetup{font={small}}
    \caption{The odometry and mapping result of Coco-LIC on the sequence \textit{degenerate\_seq\_00}, where severe LiDAR degeneration happens when Livox Avia faces the ground for a while. Coco-LIC overcomes the degradation and succeeds in returning to the origin.}

    \vspace{-0.7em}
    \label{fig:coco_map}
\end{figure}

We validate the robustness of Coco-LIC in all challenging sequences provided in R3LIVE and FAST-LIVO, using Livox Avia\footnote{\url{https://www.livoxtech.com/avia}} LiDAR at 10 Hz and its internal IMU at 200 Hz, along with a camera at 15 Hz. These sequences exhibit severe degradation, such as the solid-state LiDAR with small FoV facing the ground or walls, and the camera facing textureless surfaces or capturing blurry images due to aggressive motions, as partly shown in Fig.~\ref{fig:coco_map}.
The ground truth is not provided, but the rig starts and ends at the same position. 

Table~\ref{tab:start-end} shows the start-to-end drift error of various methods. VINS-Mono fails to work in the \textit{Visual\_Challenge} sequence due to intense motions and varying illumination, which makes it fail to stably track the visual features. The \textit{LiDAR\_Degenerate} and \textit{degenerate\_}\text{xx} sequences suffer from serious degradation of LiDAR point cloud, and when Livox Avia LiDAR faces a plane, FAST-LIO2 lacks constraints in certain DoFs, leading to insufficient pose estimation. R3LIVE and FAST-LIVO can handle various degenerations and have low start-to-end drift in most sequences, but there are also cases where both methods experience significant drift. 
In contrast, by tightly coupling LIC data, Coco-LIC achieves plausible performance in all sequences and minimal drift in several sequences. Besides, it is worth noting that although CLIC fuses all information at a time, it still fails to work in LiDAR-degenerated scenarios. This is because CLIC enforces the majority of control points in the visual sliding window to be fixed, resulting in no full exploitation of visual data~\cite{lv2023continuous}.

\subsubsection{Large-scale Dataset}
To quantitatively evaluate our approach, we conduct experiments on the large-scale UrbanNav dataset~\cite{hsu2021urbannav}. The dataset includes a 32-beam 3D LiDAR Velodyne HDL-32E\footnote{\url{https://velodynelidar.com/products/hdl-32e}} at 10 Hz, a stereo camera at 15 Hz (only use the left camera in this experiment), and an Xsens MTi-10 IMU at 400 Hz, collected in urban areas with many dynamic objects by a human-driving vehicle. 

Tab.~\ref{tab:lvi-ape} shows the RMSE of APE of the aforementioned methods. VINS-Mono performs poorly due to incorrect feature tracking caused by moving objects. LIO and LICO achieve satisfactory results in the absence of degraded LiDAR scenes. However, FAST-LIVO has slightly higher trajectory errors, possibly due to its sparse direct visual alignment being affected by moving objects and lighting changes.
Notably, in highly-dynamic motion scenarios, the accuracy of CLIC and Coco-LIC is higher attributed to their continuous-time trajectory representation, which effectively handles LiDAR distortion and efficiently fuses high-rate IMU data. Additionally, Coco-LIC reuses high-quality map points for visual factors, resulting in better performance than CLIC.

We further investigate the time consumption of competitive methods and Coco-LIC on the sequence \textit{Medium} with a duration of 785 seconds. Tab.~\ref{tab:time-cost} summarizes the result. \textit{LiDAR Association} refers to updating the local LiDAR map (global LiDAR map in the case of FAST-LIO2) and finding associations for LiDAR surf points. For CLIC, \textit{Visual Association} means visual feature extraction and tracking, and for Coco-LIC, it means updating the global LiDAR map and associating map points with the image frames. \textit{Optimization} represents executing ESIKF update or factor graph optimization. FAST-LIO2 shows brilliant efficiency due to the adoption of ESIKF and ikd-Tree~\cite{xu2022fast}. Compared to the closest work, CLIC, the proposed Coco-LIC is involved fewer control points and excludes the depth estimation process, which leads to less consuming time. Overall, all three compared methods are able to achieve real-time performance. Coco-LIC consumes around 639 seconds on the entire sequence. Note that the current implementation of Coco-LIC can be further optimized for efficiency, such as by refining the map management strategy to accelerate the association.

\begin{table}[t]
	\centering
	\captionsetup{font={small}}
	\caption{The average time consumption (milliseconds) of different modules of FAST-LIO2, CLIC, Coco-LIC on the sequence \textit{Medium}.}
 \resizebox{0.7\linewidth}{!}{
	\begin{tabular}{@{}cccc@{}}
\toprule
\multicolumn{1}{l}{} & FAST-LIO2 & CLIC & Coco-LIC \\ \midrule
LiDAR Association    & 24.14 & 32.90            & 31.46      \\
Visual Association   & - & 0.74            & 18.90      \\
Optimization         & 1.07 & 29.05            & 9.09      \\
\bottomrule
\end{tabular}}
	\label{tab:time-cost}
	\vspace{-1em}
\end{table}

\section{Conclusions and Future Work}
This letter presents Coco-LIC, a continuous-time LiDAR-Inertial-Camera odometry that tightly integrates information from LiDAR, IMU, and camera using non-uniform B-splines. The method achieves higher accuracy in pose estimation with moderate time consumption by placing more control points where motion is aggressive and fewer control points where motion is smooth. Additionally, it relies on the LiDAR map points for formulating frame-to-map visual factors. This eliminates the need for depth estimation and optimization by multiple visual keyframes, which benefits the efficiency of our continuous-time estimator.
Real-world dataset experiments demonstrate the importance of non-uniform control point placement and the effectiveness of our non-uniform continuous-time method. Robustness and accuracy evaluations show that Coco-LIC outperforms other state-of-the-art LIC Odometry, even in severe degenerate scenarios. In the future, it is worthwhile investigating  more efficient  map management strategies and incorporating complementary
 sensors like event cameras.


%

\ifCLASSOPTIONcaptionsoff
  \newpage
\fi



%

{
\AtNextBibliography{\scriptsize}
\printbibliography
}

%

\end{document}